\documentclass[11pt,a4paper]{article}
\usepackage[hyperref]{acl2020}

\usepackage{times}
\usepackage{latexsym}
\usepackage{graphicx}
\usepackage{url}
\usepackage{amsmath}
\usepackage{booktabs}
\usepackage{cleveref}
\usepackage{microtype}

\aclfinalcopy 

\title{RoNGBa: A Robustly Optimized Natural Gradient Boosting Training Approach with Leaf Number Clipping}

\author{Liliang Ren, Gen Sun and Jiaman Wu \\
 The University of California, San Diego \\
  La Jolla, CA92093 \\
   {\tt \{lren, gensun, j4wu\}@ucsd.edu} \\}
  
\date{}

\begin{document}
\maketitle
\begin{abstract}
Natural gradient has been recently introduced to the field of boosting to enable the generic probabilistic predication capability. Natural gradient boosting shows promising performance improvements on small datasets due to better training dynamics, but it suffers from slow training speed overhead especially for large datasets. We present a replication study of NGBoost \citep{NGBoost} training that carefully examines the impacts of key hyper-parameters under the circumstance of best-first decision tree learning. We find that with the regularization of leaf number clipping, the performance of NGBoost can be largely improved via a better choice of hyperparameters. Experiments show that our approach significantly beats the state-of-the-art performance on various kinds of datasets from the UCI Machine
Learning Repository while still has up to 4.85x speed up compared with the original approach of NGBoost.

\end{abstract}

\section{Introduction}
In the last decade, boosting techniques, which combine weak learners to a strong learner, have been widely developed and employed from the machine learning to computational learning communities. AdaBoost \citep{AdaBoost} and gradient boosting decision trees (GBDT) \citep{GBDT}, are some of the most popular learning algorithms used in practice. There are several highly optimized implementations of boosting, among which XGBoost \citep{XGBoost} and LightGBM \citep{LightGBM} are broadly applied to increase the scalability and decrease the complexity. These implementations can train models with hundreds of trees using millions of training examples in a matter of minutes. NGBoost \citep{NGBoost} generalized Natural Gradient as the direction of the steepest ascent in Riemannian space, and applied it for boosting to enable the probabilistic predication capability for the regression tasks. Natural gradient boosting shows promising performance improvements on small datasets due to better training dynamics, but it suffers from slow training speed overhead especially for large datasets. To reduce the training time, we consider the setting of the best-first decision tree learning \cite{best} for the weak learners, remove the restriction of maximum depth for base learners and carefully tunes the following three hyper-parameters: learning rate, number of estimators and the maximum number of leaves. Our best setting achieves up to 4.85x speed up, significantly improves the original NGBoost performance and beats
 the state-of-the-art performances on the Energy, Power and Protein datasets from the UCI Machine Learning Repository. 

\begin{table*}[t!]
\begin{center}
\begin{tabular}{lccccccc}
\toprule \bf Dataset & N & \multicolumn{2}{c}{\bf RMSE } & \multicolumn{2}{c}{\bf NLL }  &\multicolumn{2}{c}{\bf ATT }\\
&   & \bf NGBoost & \bf RoNGBa  & \bf NGBoost & \bf RoNGBa & \bf NGBoost & \bf RoNGBa    \\ \midrule
Boston   &  506 &  \bf 2.96 $\pm$ 0.42 & \bf 3.01 $\pm$ 0.57 & \bf 2.47 $\pm$ 0.12 & \bf 2.48 $\pm$ 0.16 &26.81s&\bf 10.04s\\
Concrete  & 1030 &  5.49 $\pm~0.54$ & \bf 4.71 $\pm$ 0.61 & \bf 3.08 $\pm$ 0.12 & \bf 2.94 $\pm$ 0.18 &29.96s& \bf9.28s\\
Energy & 768 & 0.51 $\pm~0.05$ & \bf 0.35 $\pm$ 0.07 & 0.76 $\pm~0.48$ & \bf 0.37 $\pm$ 0.28 & 30.24s &\bf 6.24s\\
Kin8nm  & 8192 &  0.18 $\pm~0.00$ & \bf 0.14 $\pm$ 0.00 & -0.40 $\pm~0.02$ & \bf -0.60 $\pm$ 0.03 & 189.28s &\bf 82.14s\\
Naval & 11934& \bf 0.00 $\pm$ 0.00 & \bf 0.00 $\pm$ 0.00 & -4.88 $\pm~0.04$ & \bf -5.49 $\pm$ 0.04 & 317.85s &\bf 207.01s\\
Power & 9568 &  3.92 $\pm~0.15$ & \bf 3.47 $\pm$ 0.19 & 2.80 $\pm~0.11 $ & \bf 2.65 $\pm$ 0.08 &120.31s&\bf 48.09s\\
Protein & 45730 &  4.59 $\pm~0.07$ & \bf 4.21 $\pm$ 0.06 & 2.86 $\pm~0.03$ & \bf 2.76 $\pm$ 0.03 &1191.02s&\bf 502.34s\\
Wine & 1588 &  0.64 $\pm~0.04$ & \bf 0.62 $\pm$ 0.05  &  0.94 $\pm$ 0.07  & \bf 0.91 $\pm$ 0.08 &42.44s &\bf 16.86s\\
Yacht & 308 &  \bf 0.63 $\pm$ 0.19 &  0.90 $\pm$ 0.35 & \bf 0.46 $\pm$ 0.28 & 1.03 $\pm$ 0.44 &22.52s &\bf 5.11s\\
Year MSD & 515345 &  9.18 $\pm$ NA & \bf 9.14 $\pm$ NA &  3.47 $\pm$ NA & \bf 3.46 $\pm$ NA &14.00h &\bf 5.15h\\
\bottomrule
\end{tabular}
\end{center}
\caption{\label{c} Comparison of performance between our approach (RoNGBa) and NGBoost on regression benchmark UCI datasets, where ATT means the Average Training Time. For a fair comparison, we re-run the official code of NGBoost with the hyperparameter settings reported in the paper. RoNGBa achieves significantly better results on most of the datasets apart from the extremely small (Yacht, Boston) datasets, which need extra hyperparemter tuning for better performance. 
}

\end{table*} 
\section{Robustly Optimized Natural Gradient Boosting}

Since when the maximum number of leaves is fixed, the leaf-wise tree growth algorithms (best-first) tend to achieve lower loss than the level-wise algorithms\cite{best,LightGBM}, we remove the maximum depth restriction and instead use the maximum number of leaves restriction as the regularization to prevent over-fitting. Apart from the performance gains, this change also leads to around 30\% speed up. This is because with maximum number of leaves restriction, the decision trees can often achieve lower loss by going deeper with less splits, while the decision trees bounded by maximum depth will often keep doing less effective splitting at the shallow levels.

For hyperparameter tuning, our insight is that we can counter the performance drop from decreasing the number of the weak estimators by increasing the model complexity of each base learner. In this way, the training time can be linearly reduced due to less number of weak learners for training. Since we reduce the number of weak learners and thus decrease the parameters in the system, we increase the learning rate accordingly for robust training dynamics. Based on this insight, we gradually decrease the number of estimators, while at the same time increase the maximum number of leaves and the learning rate to find the settings with the best performance. We first search for the best setting on the Energy dataset from the UCI Machine Learning Repository, and then report the performance on all datasets with the setting discovered. Generally, we use the following hyperparameters through out our experiments: learning rate, $\eta = 0.04$, number of estimators, $m=500$, maximum number of leaves, $n=31$.

\section{Experiments}

Our experiments use datasets from the UCI Machine
Learning Repository, and follow the same protocol as NGBoost \cite{probabilistic,NGBoost}. For
all datasets, we hold out a random 10\% of the examples as a test set. From the other 90\% we initially hold
out 20\% as a validation set to select $M$ (the number of
boosting stages) that gives the best log-likelihood, and
then re-fit the entire 90\% using the chosen $M$. The refit model is then made to predict on the held-out 10\%
test set. This entire process is repeated 20 times for
all datasets except Protein and Year MSD, for which
it is repeated 5 times and 1 time respectively. For the Average Training Time (ATT) measurement, we take an average of the training times measured from each of the repeated training processes. Unlike the original implementation, we use the learning rate of 0.04 throughout all the datasets.

We also re-run the official NGBoost code with the same hyper-parameters as claimed in the original paper for a fair comparison of the performance and the training time. All the experiments are conducted on a single Intel(R) Xeon(R) E5-2630 v4 2.20GHz CPU. 

\section{Results}
\Cref{c} compares the performance of our approach with the original approach of NGBoost on the regression benchmark of UCI datasets. We can see that RoNGBa achieves significantly better results on most of the datasets apart from extremely small (Yacht, Boston) datasets, which need extra hyperparemter tuning for better performance. Specifically, our approach significantly beat the state-of-the-art performances on the Energy, Power and Protein datasets as reported from \citet{dropout} and \citet{simple}. We can also observe that our approach can achieve a speed up ranging from 1.53x to 4.85x in various kinds of datasets, which empirically confirms our insight that reducing the overall number of learners can cut down much more amount of computation time than the time gained from increasing each base learner's model complexity.

\section{Related Work}

\noindent \textbf{AdaBoost} \cite{AdaBoost} changes the input distribution to obtain subsequent answers from the former weak learners. At each training step, it puts higher weights on mis-classified examples, and finally composes a strong classifier by weighted sum of all the weak hypotheses.

\noindent \textbf{Gradient Boosting Decision Tree (GBDT)} \cite{AdaBoost} is adapted from Adaboost in order to handle a variety of loss functions. GBDT first expresses the loss function minimization problem into an additive model, and performs numerical optimization directly in the function space applying greedy forward stage-wise algorithm. Most importantly, GBDT uses the data-based analogue of the unconstrained negative gradient of the loss function in the current model as the approximate value of the residual in boosting tree, which gives the best steepest-descent step direction in the N-dimensional data space.

Compared with AdaBoost, GBDT constructs multiple decision trees serially to predict the data. It takes the decision tree model as parameter and each iteration is fitted to the negative gradient of the loss function to improve. However, AdaBoost takes each point as parameter and adjusts the weight of the negative points to improve. Therefore, by choosing different types of loss functions , such as square error and absolute error in regression, negative binomial log-likelihood error in classification, GBDT can be applied to broader and more diverse learning problems than AdaBoost, like multi-class classification, click prediction, and learning to rank.
 
\noindent \textbf{XGBoost} \citep{XGBoost} improves GBDT with better scalability. XGBoost is suitable for large scale data and limited computing resource with high speed and equivalent accuracy. To achieve this scalability, XGBoost uses mainly three techniques to improve: 1) XGBoost approximates the best split of decision trees by weighted quantile sketch, instead of greedily computing all possible splits. 2) XGBoost handles sparse data by sparsity-aware algorithm which only trains non-missed data and gets a default tree direction for missing values. 3) XGBoost stores memory with a cache-aware block structure for out-of-core computing.

\noindent \textbf{LightGBM} \cite{LightGBM} further improves the system scalability for high-dimensional large data. They apply two methods, Gradient-based One-Side Sampling (GOSS) and Exclusive Feature Bundling (EFB) on GBDT to increase the efficiency without hurting the accuracy. 

GOSS samples training data by keeping all the instances with large gradients and random sampling on the instances with small gradients since instances with small gradients are already well-trained. Then, to keep data distribution, they amplify the sampled data with small gradients via a constant during computing the information gain. Instead of filtering out data with zero values as training data in XGBoost, LightGBM samples the training dataset more wisely.  

In reality, there are features mutually exclusive and thus data can be very sparse. To reduce the number of features, EFB bundles the exclusive features into a single feature. First, they take features as vertices and add edges between not mutually exclusive features. Edges are weighted by total conflicts between features. Then, they sort the features by degrees in the graph. Finally, they put a feature in the sorted list to  an existing bundle or a new created one based on the conflicts comparing to a threshold. After feature histograms are constructed, they find the best split points by histogram-based algorithm, comparing to XGBoost approximates the best split points by weighted quantile sketch. 

Even though LightGBM does not apply new techniques, such as cache-aware blocks and out-of-core computing in XGBoost, to interact with system more efficiently, LightGBM still outperforms XGBoost with more efficient algorithm.

\section{Conclusion}
In this work, we proposed \textbf{RoNGBa}, a \textbf{R}obustly \textbf{o}ptimized \textbf{NGB}oost \textbf{a}pproach. RoNGBa applies leaf number clipping for base learners and find the best hyperparameters based on a simple yet effective insight on computation-accuracy trade-off. Our approach significantly beats the state-of-the-art performance on various kinds of UCI datasets while still has up to 4.85x speed up compared with the original approach of NGBoost.

Our future work is to apply the techniques of Gradient-based One-Side Sampling and Exclusive Feature Bundling from LightGBM for more efficient natural gradient boosting on large-scale higher-dimensional datasets.

\section*{Acknowledgments}

We want to thank Michal Moshkovitz and Joseph Geumlek for the early discussions of the project. 

\bibliographystyle{acl_natbib}
\bibliography{acl2020} 

\appendix

\end{document}